\documentclass[sigconf]{acmart}

\renewcommand\footnotetextcopyrightpermission[1]{} 

\setcopyright{NONE}

\AtBeginDocument{%
  }

\acmSubmissionID{1010}

\usepackage{subcaption}
\usepackage{graphbox}
\usepackage{wrapfig}
\usepackage{comment}
\usepackage{booktabs}
\usepackage{multirow}

\begin{document}
\bibliographystyle{ACM-Reference-Format}
\title[Hand-Object Interaction Controller (HOIC)]{Hand-Object Interaction Controller (HOIC): Deep Reinforcement Learning for Reconstructing Interactions with Physics}
\author{Haoyu Hu}
\email{hhythu17@163.com}
\orcid{0000-0002-4273-1245}
\affiliation{%
  \institution{School of Software and BNRist, Tsinghua University}
  \city{Beijing}
  \country{China}
}

\author{Xinyu Yi}
\email{yixy20@mails.tsinghua.edu.cn}
\orcid{0000-0003-3504-3222}
\affiliation{%
  \institution{School of Software and BNRist, Tsinghua University}
  \city{Beijing}
  \country{China}
}

\author{Zhe Cao}
\email{zhecao@google.com}
\orcid{0000-0002-8704-473X}
\affiliation{%
  \institution{Google}
  \city{Seattle}
  \country{United States of America}
}

\author{Jun-Hai Yong}
\email{yongjh@tsinghua.edu.cn}
\orcid{0000-0002-4326-4167}
\affiliation{%
  \institution{School of Software and BNRist, Tsinghua University}
  \city{Beijing}
  \country{China}
}

\author{Feng Xu}
\authornote{Corresponding author.}
\email{xufeng2003@gmail.com}
\orcid{0000-0002-0953-1057}
\affiliation{%
  \institution{School of Software and BNRist, Tsinghua University}
  \city{Beijing}
  \country{China}
}

\renewcommand{\shortauthors}{Hu et al.}

\begin{abstract}
Hand manipulating objects is an important interaction motion in our daily activities.
We faithfully reconstruct this motion with a single RGBD camera by a novel deep reinforcement learning method to leverage physics. 
Firstly, we propose object compensation control which establishes direct object control to make the network training more stable.
Meanwhile, by leveraging the compensation force and torque, we seamlessly upgrade the simple point contact model to a more physical-plausible surface contact model, further improving the reconstruction accuracy and physical correctness. 
Experiments indicate that without involving any heuristic physical rules, this work still successfully involves physics in the reconstruction of hand-object interactions which are complex motions hard to imitate with deep reinforcement learning.
Our code and data are available at \url{https://github.com/hu-hy17/HOIC}.
\end{abstract}



\begin{teaserfigure}
    \centering
    \includegraphics[width=0.95\linewidth]{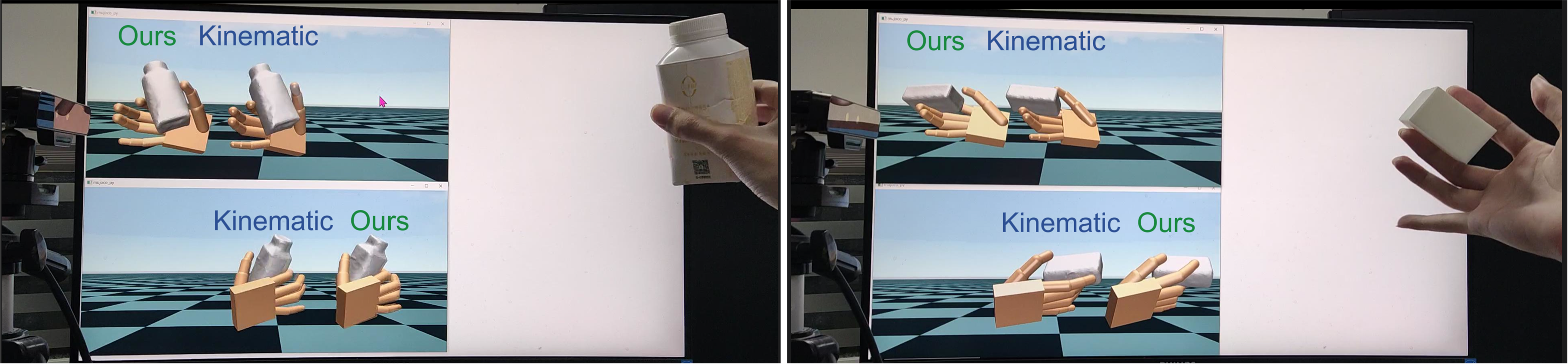}
    \caption{Our HOIC framework reconstructs accurate and physically plausible hand-object interaction motions by imitating the vision-based kinematic tracking results in the physics simulator.}
    \label{fig:teaser}
\end{teaserfigure}

\maketitle

\section{Introduction}
%
Our hands play a crucial role in our interactions with the world around us. 
Whether we're shopping and selecting items from shelves or preparing food by skillfully wielding a knife to cut ingredients into small pieces, our hands are at the forefront of these actions.
The accurate and real-time reconstruction of such Hand-Object Interaction (HOI) motions can facilitate a variety of applications including virtual reality, human-computer interaction, and robot learning.
\par
Many works have focused on the challenging task of jointly reconstructing hand pose and object motion from monocular RGB or depth input only. 
To deal with the ill-posed nature of this problem, some recent studies utilize physical laws to regularize or refine the reconstruction procedures, which can leverage both data-driven and model-driven methodologies to yield more accurate and physically plausible results \cite{wang2013video, zhao2022stability, hu2022physical, tzionas2016capturing}. 
However, these methods either perform time-consuming optimization processes that are not suitable for real-time applications~\cite{wang2013video, zhao2022stability, tzionas2016capturing} or exert a strong restriction in the scheme when performing the physical refinement~\cite{hu2022physical}.
A more powerful solution to incorporate physics in the reconstruction of HOI is eagerly required.

%
Imitation learning techniques are proven to be successful in generating physically plausible results and have been widely used in human motion synthesis \cite{peng2018deepmimic, yuan2020residual, peng2021amp} and capture \cite{yuan2021simpoe, fussell2021supertrack} in recent years.
These techniques usually train an imitation policy network under the reinforcement learning framework.
The policy network takes reference non-physical motions as input and output control signals to generate physically plausible motions.
However, applying these techniques in the field of HOI reconstruction still faces some challenges:
%
%
Firstly, a HOI controller needs to accomplish a more complex task than that of a human body controller.
Under HOI scenarios, the objects can only be indirectly controlled through the contact force exerted by the hand. 
The HOI controller not only has to mimic the reference hand motion but also needs to generate accurate contact points and forces to drive the manipulated object and ensure the object follows the reference object motion, which is very challenging to learn.
Secondly, the contact mechanism in modern physics simulators differs significantly from the HOI contact in the real physical world. 
In the real world, human hands have soft tissues that can deform and generate contact areas when contacting an object.
In traditional simulators, however, an approximation is made using contact points.
This limitation prevents many real interactive actions from being imitated in the physical simulation.
\par
In this paper, we propose HOIC (Hand-object Interaction Controller), a deep reinforcement learning (DRL) method for imitating and reconstructing HOI motion, which effectively incorporates physics and addresses the aforementioned challenges. 
%
%
%
We introduce an innovative mechanism called object compensation control, designed to address the discrepancies between the physical simulator's contact representation and the real-world contact mechanism.
In addition to hand control signals, our policy network generates supplementary force and torque, directly influencing the object during simulation. This direct control significantly improves system stability in the imitation task.
The forces and torques, referred to as compensation force and torque, could be well explained by a surface contact model. This model implicitly allows the system to simulate surface contacts instead of point contacts during hand-object interactions.
In this manner, we avoid the task of modeling the soft tissue surface contact in the forward physical model, which is known as a complex and time-consuming process. 
Finally, the unexplained compensation force and torque are formed as the penalty of a reward term during policy training.
The proposed object compensation control not only simplifies the HOI imitation task but also enhances its physical plausibility.
The approach ensures a more direct and realistic interaction between the hand and the object, contributing to the overall effectiveness of the imitation process.
%
%
\par
In the experiments, we demonstrate that object compensation control can significantly improve training speed and imitation quality.
We also incorporate the imitation policy into a real-time HOI tracking system and prove that it can reconstruct more physically plausible HOI motions compared with the pure vision-based method \cite{zhang2021single} and the method with heuristic physical refinement \cite{hu2022physical}.

\section{Related Works}
As our work focuses on reconstructing physically plausible HOI motion via the motion imitation controller, we will review recent works on HOI reconstruction and physics-aware human motion control.

\subsection{Hand-object Interaction Reconstruction}
Joint reconstruction of interacting hand object motions has been studied for many years. 
The heavy occlusion between hand and object during interaction makes this task challenging and ill-posed. 
Earlier works use multi-view camera input to minimize the impact of occlusion \cite{oikonomidis2011full, ballan2012motion}. 
With the development of depth cameras, some works \cite{kyriazis2013physically, schmidt2015depth, sridhar2016real, tsoli2018joint, zhang2021single, chen2023tracking} utilize RGBD images as input to obtain more 3D cues, improving reconstruction precision. 
In recent years, many HOI datasets \cite{FirstPersonAction_CVPR2018, hampali2020honnotate, Brahmbhatt_2020_ECCV, chao:cvpr2021, fan2023arctic} have been released, which encourage the community to leverage data-driven methods to reduce the ambiguity of hand object pose estimation \cite{hasson2019learning, tekin2019h+, karunratanakul2020grasping, chen2023gsdf}.

However, these works still suffer from some noticeable, especially physically implausible artifacts including jitter, penetration, and unstable manipulate pose.
Some studies introduce physical priors to regularize the estimated hand object pose. 
\citet{wang2013video} leverage physical simulation and trajectory optimization to find an optimal motion control that best matches the input data. 
\citet{tzionas2016capturing} and \citet{zhao2022stability} optimize hand pose in each frame to form a stable object grasping. 
However, the optimization approaches used in these works are either offline or time-consuming, which hinders their usage in real-time applications.
\citet{hu2022physical} utilizes object dynamics and a simplified frictional contact model to correct hand pose to match object movement, which can run in real-time.
However, it assumes that contact only happens on fingertips, thus being unable to handle general HOI scenarios.
Our method overcomes these shortcomings by leveraging the imitation learning technique where the physical prior is learned by mimicking various interaction motions.

\subsection{Physics-Aware Human Motion Control}
\subsubsection{Optimization-based Method}
Many works on human motion incorporate physical laws using optimization-based methods.
Some works achieve physically plausible body motion capture by optimizing joint torques and foot contact forces to drive dynamic simulation \cite{zheng2013human, PhysCap, xie2021physics, PIP}.
In the realm of HOI, the dynamics of hands and objects, and the physical mechanisms of contact are studied and widely applied in motion synthesis tasks \cite{kry2006interaction, ye2012synthesis, mordatch2012contact, liu2009dextrous, wu2023learning}.
For example, \citet{liu2009dextrous} generates various motions in response to different dynamic situations during manipulation by leveraging the physical relationship between object motion and hand actuation torques.
\citet{wu2023learning} utilizes a bilevel optimization to synthesize diverse grasping motions on different objects while maintaining physical correctness.
However, due to the complex contact mechanisms in HOI, these optimization methods are either time-consuming or designed only for grasping tasks. 
Therefore, our work aims to construct a more efficient and general physical optimization method in HOI using DRL.

\subsubsection{DRL-based Method}
In recent years, deep reinforcement learning (DRL) has emerged as a powerful approach for controlling physically simulated characters, yielding impressive results. 
\citet{peng2018deepmimic} learns a control policy by imitating example data, which can generate high-quality, physically plausible human motions.
\citet{yuan2020residual} improves the ability of the mimic policy by introducing residual force control. 
Instead of imitating a single motion at once, some works \cite{peng2021amp, peng2022ase, tessler2023calm} utilize GAIL \cite{ho2016generative} to learn stylized motion skills from multiple motion clips, which can generate diverse movements by combining different skills. 
Besides motion synthesis, some works utilize DRL for motion capture.
\citet{yuan2021simpoe} trains an imitation control policy to follow the reference human motion captured by the vision-based method. 
\citet{winkler2022questsim} uses DRL to learn the mapping between sparse signals from VR devices and full-body motion.
By incorporating modern physical simulators, these methods can yield physically plausible results when reconstructing real-world human motion.

DRL is also widely used in HOI control for robot learning and motion synthesis.
\citet{andrychowicz2020learning} and \citet{chen2022system} train hand controllers to finish object reorientation tasks on robotic hands.
\citet{chen2022towards} explores several manipulation tasks involving two dexterous hands.
Some works use DRL to learn complex interaction skills, such as using chopsticks \cite{yang2022learning}, playing basketball \cite{liu2018learning}, and pre-grasping \cite{chen2023synthesizing}.
Imitation learning has gained popularity in recent years, aiming to enhance sample efficiency during training and the naturalness of manipulation motions yielded by the policy.
\citet{sermanet2018time} and \citet{liu2018imitation} improve the manipulation ability of robots by imitating expert motions extracted from raw videos. 
However, these works employ two-finger grippers which enable fewer interactive movements compared to the human hand.
\citet{rajeswaran2017learning}, \citet{radosavovic2021state}, and \citet{qin2022dexmv} perform imitation learning using a dexterous hand model, generating manipulation motions that are close to the real human hand.
However, all these works focus on finishing a specific manipulation task such as moving, pouring, reorientation, etc. 
Our goal is to develop a more general policy that can be used to imitate any given human-object interaction.

\begin{figure*}[!t]
    \centering
    \includegraphics[width=0.95\linewidth]{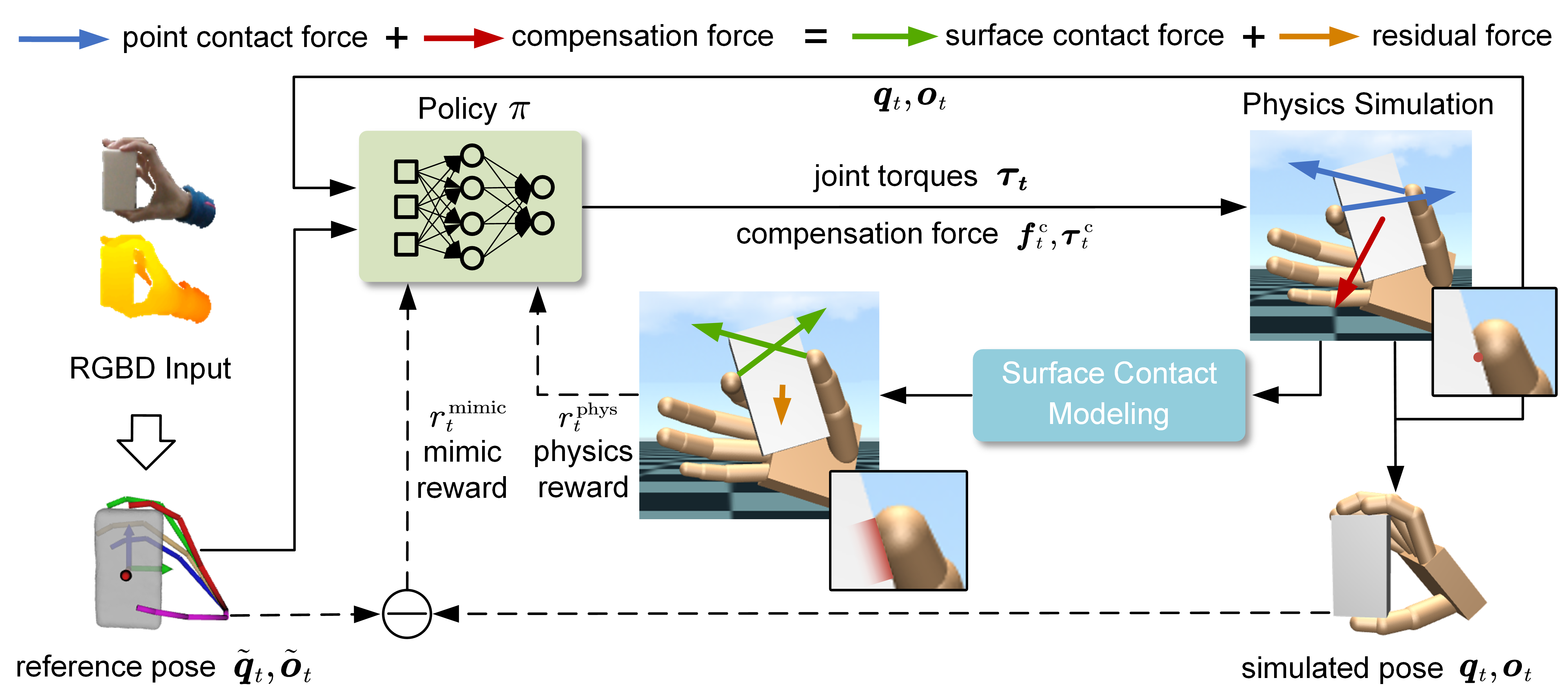}
    \caption{System Overview. Our system takes the kinematic tracking results of hand-object interaction $\{\boldsymbol{\tilde{q}}_{t}, \boldsymbol{\tilde{o}}_{t}\}$ as input and outputs a refined version $\{\boldsymbol{q}_{t}, \boldsymbol{o}_{t}\}$. The policy network $\pi$ first takes $\{\boldsymbol{q}_{t}, \boldsymbol{o}_{t}\}$ and $\{\boldsymbol{\tilde{q}}_{t+i}, \boldsymbol{\tilde{o}}_{t+i}\}$ as input to predict the control signals $\{\boldsymbol{\tau}_t, \boldsymbol{f}_t^c, \boldsymbol{\tau}_t^c\}$ which are fed into a physical simulator to obtain $\{\boldsymbol{q}_{t+1}, \boldsymbol{o}_{t+1}\}$. In the training process, the compensation force $\{\boldsymbol{f}_t^c, \boldsymbol{\tau}_t^c\}$ is applied to upgrade the contact model from point to surface contact in a Surface Contact Modeling step and the residual is used to construct the physics reward $r_t^{\text{phys}}$. Meanwhile, a mimic reward $r_t^{\text{mimic}}$ is also used to train the policy network $\pi$.}
    \label{fig:pipeline}
\end{figure*}

\section{Method}
We aim to reconstruct hand-object interaction motion from a single-view camera in real-time. 
Our method takes a monocular RGBD video stream as input and outputs the hand pose and the object motion of each frame (Fig.~\ref{fig:pipeline}).
Firstly, we use an off-the-shelf kinematic tracking method~\cite{zhang2021single} to obtain a coarse estimation of hand pose $\boldsymbol{\tilde{q}}_t$ and object pose $\boldsymbol{\tilde{o}}_t$.
Without careful physics-aware modeling, the kinematic interaction motion may contain many physically implausible artifacts such as penetration, jitter, missing contact points, and non-physical sliding.
Then the kinematic results are used as reference motions and an imitation process is performed to refine the kinematic results (Sec.~\ref{sec:imitation}).
To be specific, at each time step $t$ of the imitation, a policy $\pi$ takes the refined result $\{\boldsymbol{q}_{t},\boldsymbol{o}_{t}\}$ at $t$ and several future frames of the kinematic results $\{\boldsymbol{\tilde{q}}_{t+1},\boldsymbol{\tilde{o}}_{t+1},...,\boldsymbol{\tilde{q}}_{t+N},\boldsymbol{\tilde{o}}_{t+N}\}$ as input.
%
The policy outputs control signals, including not only the joint torques $\boldsymbol{\tau}_t$ for the hand but also the compensation force $\boldsymbol{f}_t^c$ and torque $\boldsymbol{\tau}_t^c$ for the object, which are all fed into a physical simulator to simulate physical-plausible results.
To train the policy, we utilize two rewards: the mimic reward $r_t^{\text{mimic}}$, and the physics reward $r_t^{\text{phys}}$.

The mimic reward is a standard imitation learning metric that encourages the policy to replicate the reference motion. 
The physics reward acts as a constraint, ensuring that the training is stable while the results adhere to physical laws.
More specifically, to compute $r_t^{\text{phys}}$, we use a surface contact model to explain the compensation force (force and torque, using force for simplification) for more physical correctness (Sec.~\ref{sec:surface}). 
While the surface contact model can explain most of the compensation force, a portion remains unaccounted for, which is referred to as the residual force.
The physics reward guides the policy towards physically realistic compensation force usage by minimizing the residual force. 
This ensures that the generated forces align with the laws of physics.

%
%
%
%
%
%
\subsection{HOI Imitation Learning}\label{sec:imitation}
To imitate the hand-object interaction with a physics simulator, we apply a standard reinforcement learning framework \cite{sutton1998introduction}.
Reinforcement learning aims to find an optimal solution for a sequential decision process.
In the process, an agent interacts with the environment according to a policy $\pi$ which is typically represented as a conditional distribution.
At each time step of the sequence, the agent observes the state $s_t$ of the environment and takes an action $a_t$ sampled from the distribution $\pi(a|s)$.
After taking the action, the agent receives a reward $r(s_t, a_t)$ and the new state $s_{t+ 1}$.
The optimization objective is to find the optimal distribution $\pi(a|s)$ that maximizes the accumulated rewards over the entire sequence, written as:
\begin{equation}\label{eq:totalrwd}
   \arg\max\limits_{\pi} E_{\tau \sim p_{\pi}(\tau)}\left(\sum_{t}r(s_t, a_t)\right),
\end{equation}
where $\tau$ stands for the whole state and action sequence $[s_0, a_0, s_1, a_1,\\ ..., s_T, a_T]$, and $p_{\pi}(\tau)$ is the distribution of $\tau$ when the agent follows the policy $\pi$.
\par
Deep Reinforcement Learning (DRL) uses a neural network to represent the policy, enabling the learning of complex mappings between high-dimensional environmental states and actions.
In our work, we follow the widely used reinforcement learning algorithm Proximal Policy Optimization (PPO) \cite{schulman2017proximal} to train our policy network.
In the following part of this section, we will elaborate on the definition of state, action, and reward in our method.


\subsubsection{State}
The state is defined as:
\begin{equation}\label{eq:state}
\begin{split}
    \boldsymbol{s}_t=(\boldsymbol{q}_t, 
    \boldsymbol{\dot{q}}_t,    
    &\boldsymbol{o}_t, 
    \boldsymbol{\dot{o}}_t,
    \boldsymbol{\tilde{q}}_{t+i}, 
    \boldsymbol{\tilde{o}}_{t+i}
    ), 
    \\
    &i=1...N,
\end{split}
\end{equation}
which contains not only the hand pose $\boldsymbol{q}_t$, the hand velocity $\boldsymbol{\dot{q}}_t$, the object pose $\boldsymbol{o}_t$, and the object velocity $\boldsymbol{\dot{o}}_t$ at the current frame, but also the kinematic hand pose $\boldsymbol{\tilde{q}}_{t+i}$ and the kinematic object pose $\boldsymbol{\tilde{o}}_{t+i}$ in the future frames.
Considering that the action should be independent of the absolute positions of the hands and objects, we do not include the current hand root translation in the state and transform all the components in Eq.~\ref{eq:state}, except hand root rotation contained in $\boldsymbol{q}_t$, to the current hand root coordinate.

\subsubsection{Action}
The action $\boldsymbol{a}_t$ contains control signals that are used to drive the proxy hand and object in the physics simulator.
%
%
For hand control, instead of directly outputting joint torques from the policy, we first output a target hand pose $\hat{\boldsymbol{q}}_t$ and employ a PD-controller to solve for the joint torques following previous approaches \cite{yuan2021simpoe, christen2022d}.
To be specific, the PD-controller takes the current hand pose $\boldsymbol{q}_t$, velocity $\boldsymbol{\dot{q}}_t$, and the target hand pose $\boldsymbol{\hat{q}}_t$ as input.
It outputs joint torques $\boldsymbol{\tau}_t$ that drive the hand to move towards the target pose, written as:
\begin{equation}
\boldsymbol{\tau}_t = \boldsymbol{k}_p \circ (\boldsymbol{\hat{q}}_t - \boldsymbol{q}_t) - \boldsymbol{k}_d \circ \boldsymbol{\dot{q}}_t,
\end{equation}
where $\circ$ denotes vector element-wise multiplication, $\boldsymbol{k}_p$ and $\boldsymbol{k}_d$ are the parameters of the PD-controller.
\par
Different from previous works on hand manipulation control, we augment our action space with a compensation force $\boldsymbol{f}_t^c, \boldsymbol{\tau}_t^c$ that is directly exerted on the object.
Physically, the object is driven by the hand and the compensation force does not exist.
However, in the reinforcement learning framework, it is difficult to train the policy for the hand joints with rewards defined on the manipulated object, as the control is quite indirect.
Involving this compensation force makes the training more stable and thus enables our policy to track a wider range of interaction motions.

%
Finally, our action is defined as:
$\boldsymbol{a}_t = \{\boldsymbol{\tau}_t, \boldsymbol{f}_t^c, \boldsymbol{\tau}_t^c\}$.
Note that the problem of the physical incorrectness involved by the compensation force will be handled later.

\subsubsection{Reward}
The goal of our policy is to imitate a given hand-object interaction motion while using a physically valid compensation force to control the object. 
Therefore, the reward at each time step is defined as the multiplication of two types of reward: mimic reward $r_t^{\text{mimic}}$ and physics reward $r_t^{\text{phys}}$:
\begin{equation}
    r_t = r_t^{\text{mimic}} \cdot r_t^{\text{phys}}.
\end{equation}
The mimic reward $r_t^{\text{mimic}}$ encourages the policy to generate a similar interaction motion to the reference motion, which is defined as:
\begin{equation}
    r_t^{\text{mimic}} = r_t^{\text{hand}} \cdot r_t^{\text{object}},
\end{equation}
where $r_t^{\text{hand}}$ evaluates the similarity of hand motion between the imitation result and the reference. Inspired by previous work on human body imitation \cite{yuan2021simpoe}, our hand reward $r_t^{\text{hand}}$ consists of four parts:
\begin{equation}
\label{eq:hand_reward}
    r_t^{\text{hand}} = w^{\text{pose}} r_t^{\text{pose}} + w^{\text{joint}} r_t^{\text{joint}} + w^{\text{orient}} r_t^{\text{orient}} + w^{\text{vel}} r_t^{\text{vel}},
\end{equation}
where $r_t^{\text{pose}}$ measures the difference between the current hand pose $\boldsymbol{q}_{t}$ and the reference hand pose $\boldsymbol{\tilde{q}}_{t}$:
\begin{equation}
\label{eq:hand_pose}
    r_t^{\text{pose}} = \text{exp}\left(-k^{\text{pose}}||\boldsymbol{q}_{t} - \boldsymbol{\tilde{q}}_{t}||\right).
\end{equation}
$r_t^{\text{joint}}$ evaluates the difference between the current 3D hand joint positions $\boldsymbol{j}_{t}$ and the reference hand joint positions $\boldsymbol{\tilde{j}}_{t}$:
\begin{equation}
\label{eq:hand_joint}
    r_t^{\text{joint}} = \text{exp}\left(-k^{\text{joint}}||\boldsymbol{j}_{t} - \boldsymbol{\tilde{j}}_{t}||\right).
\end{equation}
$r_t^{\text{orient}}$ encourages the orientation of each bone $R_{t}^i$ of the current hand is close to that of the reference $\tilde{R}_{t}^i$, computed as:
\begin{equation}
\label{eq:hand_orient}
    r_t^{\text{orient}} = \text{exp}\left(-k^{\text{orient}}\sum_{i=1}^B|R_{t}^i \ominus \tilde{R}_{t}^i|\right),
\end{equation}
where $B$ is the number of bones and $\ominus$ denotes the relative rotation angle between the two rotation matrices. 
$r_t^{\text{vel}}$ guarantees the velocity of hand motion $\boldsymbol{\dot{q}}_t$ should be similar to the reference motion $\boldsymbol{\dot{\tilde{q}}}_{t}$:
\begin{equation}
\label{eq:hand_vel}
    r_t^{\text{vel}} = \text{exp}\left(-k^{\text{vel}}||\boldsymbol{\dot{q}}_t - \boldsymbol{\dot{\tilde{q}}}_{t}||\right),
\end{equation}
where $\boldsymbol{\dot{q}}_t$ comes from the simulator whereas $\boldsymbol{\dot{\tilde{q}}}_{t}$ is calculated using finite difference. 
Similar to $r_t^{\text{hand}}$, $r_t^{\text{object}}$ measures the object motion difference, which is computed as:
\begin{equation}
    r_t^{\text{object}} = w^{\text{opose}} r_t^{\text{opose}} + w^{\text{ovel}} r_t^{\text{ovel}},
\end{equation}
where $r_t^{\text{opose}}$ and $r_t^{\text{ovel}}$ are calculated similar to $r_t^{\text{pose}}$ in Eq.~\ref{eq:hand_pose} and $r_t^{\text{vel}}$ in Eq.~\ref{eq:hand_vel}, respectively.

%
Different from previous imitation learning methods \cite{peng2018deepmimic, yuan2021simpoe}, our method incorporates a novel physics reward $r_t^{\text{phys}}$ in addition to the imitation reward. 
This physics reward is designed to constrain the use of compensation forces.
%
While compensation forces enable the policy to imitate a wider range of hand-object interactions, they can also be exploited to maximize imitation rewards in ways that violate physical laws. 
To prevent this, we aim to ensure that the policy's compensation force remains within a physically plausible range. 
A naive solution is to restrict the compensation force to be as small as possible as it is not a real existing force.
However, we notice that the point contact model used in the physics simulator is efficient but not physically correct.
The compensation force could be used to model this incorrectness and thus does not need to be zero.
Based on this observation, we propose to use a surface contact model (detailed in Sec.~\ref{sec:surface}) to measure how much of the compensation force is modeling the surface contact and we just restrict the residue, so-called residual force, to be as small as possible.
%
%
In this manner, the physics reward $r_t^{\text{phys}}$ reflects how well the simulation adheres to real physics.

Note that all the sub-rewards, including the imitation rewards for both the hand and the object, as well as the physics reward, are multiplied together to form the final reward.
This design forces the policy to accurately imitate hand and object movements while employing compensation forces in a physically realistic manner.
%
%
%
\subsection{Surface Contact Modeling}\label{sec:surface}
In this section, we will elaborate on the computation of physics reward $r_t^{\text{phys}}$ using our surface contact model.
%
%
For brevity, the subscript for time step $t$ will be omitted in the remainder of this section.

The input of surface contact modeling consists of hand-object contact points $\{\boldsymbol{p}_i\}$, normals $\{\boldsymbol{n}_i\}$, object mass center $\boldsymbol{c}$, translation acceleration $\boldsymbol{\dot{v}}$, angular velocity $\boldsymbol{\omega}$ and acceleration $\boldsymbol{\dot{\omega}}$, mass $m$, and inertia $\mathcal{I}$.
These data can be obtained from the physics simulator.
Firstly, we expand each contact point to form a contact surface.
Some previous works implement surface contact mechanisms by using a deformable hand model \cite{jain2011controlling} or by adding torsional and rolling torques to the point contact model \cite{luding2008cohesive}.
However, the former needs to use the meshes of hand and object to compute an accurate contact surface, which is undesirable considering the high sample efficiency required by reinforcement learning. The latter needs additional physical hyperparameters to ensure accuracy.
Therefore, we consider a simple but efficient model in our method.
%
%
%
For each contact point, we sample four additional points alongside four tangential directions, and these points are 2.5 millimeters away from the original point.
These additional points, combined with the original contact points, are referred to as extended contact points.
It can be proved that on the rectangle enclosed by the additional points, any distribution of surface contact forces can be equivalently represented by a combination of point contact forces at the extended contact points \cite{caron2015stability}.
After the expansion, we obtained a set of new contact points that is five times more than the original ones, denoted as $\{\boldsymbol{p}_i^{\text{ext}}\}$.

Next, we optimize a set of forces at each extended contact point to match the dynamics of the object.
Similar to the previous work \cite{hu2022physical}, we use the Coulomb Friction Model to represent the contact force.
The contact force is composed of pressure and frictional force, and the resultant force of these two must be within the range of the friction cone.
Following \cite{lynch2017modern}, we use a quadrangular pyramid to approximate the friction cone for computational convenience.
Each contact force $\boldsymbol{f}_i$ is represented by the positive span of four normalized lateral edges $\boldsymbol x_1,\cdots,\boldsymbol x_4$ of the pyramid, written as 
$\boldsymbol{f}_i = \boldsymbol{A}_i\boldsymbol{\lambda}_i, (\boldsymbol{\lambda}_i \in \mathbb R^4, \boldsymbol{\lambda}_i \ge 0)$, 
where $\boldsymbol{A}_i=[\boldsymbol x_1,\boldsymbol x_2,\boldsymbol x_3,\boldsymbol x_4]$.
The optimization problem is defined as 
\begin{equation}
\label{eq:force_opt}
\begin{split}
    E(\boldsymbol{\lambda}_i) = E_{\text{force}} &+ E_{\text{torque}} + E_{\text{rel}} \\
    \text{s.t. } &\boldsymbol{\lambda}_i\ge0.
\end{split}
\end{equation}
$E_{\text{force}}$ requires that the sum of contact forces and the object gravity $m\boldsymbol{g}$ match the translation acceleration $\dot{\boldsymbol{v}}$ of the object:
\begin{equation}
\begin{split}
    E_{\text{force}} =\Big|\Big|\sum_{i=1}^{P} \boldsymbol{A}_i&\boldsymbol{\lambda}_i - \boldsymbol{f}_{\text{target}}\Big|\Big|^2, \\ 
    \boldsymbol{f}_{\text{target}} = m&\dot{\boldsymbol{v}} - m\boldsymbol{g},
\end{split}
\end{equation}
where $P$ is the number of extended contact points.
Similar to $E_{\text{force}}$, $E_{\text{torque}}$ evaluates the consistency between total contact torques and the rotational movement of the object:
\begin{equation}
\begin{split}
    E_{\text{torque}} =\Big|\Big|\sum_{i=1}^{P} [\boldsymbol{p}_i^{\text{ext}} & - \boldsymbol{c}]_\times \boldsymbol{A}_i\boldsymbol{\lambda}_i-\boldsymbol{\tau}_{\text{target}}\Big|\Big|^2, \\
    \boldsymbol{\tau}_{\text{target}} = \mathcal{I}&\dot{\boldsymbol{\omega}} + [\boldsymbol{\omega}]_\times \mathcal{I} \boldsymbol{\omega},
\end{split}
\end{equation}
where $\boldsymbol{c}$ is the mass center of the object, $[\cdot]_{\times}$ is the cross product matrix.
Finally, $E_{\text{rel}}$ requires that when there is relative sliding between the hand and the object at the contact point, the frictional force should act in the direction that prevents sliding:
\begin{equation}
\begin{split}
    E_{\text{rel}} =\sum_{i=1}^{P} (&\beta_i \boldsymbol{e} + \boldsymbol{A}_i^T \boldsymbol{v}_{\text{t}_i}^{\text{rel}}) \cdot \boldsymbol{\lambda}_i, \\
    \beta_i = &\|\boldsymbol{A}_i^T \boldsymbol{v}_{\text{t}_i}^{\text{rel}} \|_{\infty},
\end{split}
\end{equation}
where $\boldsymbol{e}$ is a vector of ones, $\boldsymbol{v}_{\text{t}_i}^{\text{rel}}$ is the relative tangential velocity at the contact point $\boldsymbol{p}_i^{\text{ext}}$, $\cdot$ is the dot product of two vectors, $||\cdot||_{\infty}$ is the maximum among the absolute values of the vector elements.
%
%
%
%
For more details about the meaning of $E_{\text{rel}}$, we recommend readers refer to \cite{andrews2022contact}.

The optimization problem in Eq.~\ref{eq:force_opt} can be solved using quadratic programming.
This process yields a solution comprising a set of contact forces, named net contact forces, which optimally align with the dynamics of the object.
Note that we just calculate these forces to model the surface contact but have not used them in the imitation process. 
In our system, we do not need to directly involve them in the imitation process, which is complex and time-consuming. 
The reason is that we have a compensation force that can be used to represent these forces and is already involved in the imitation process.
In this case, the residue in the compensation force, which cannot be represented by the net contact forces, can be treated as the difference between the net contact forces and the proper driving force for object motion, which is named residual force $\boldsymbol{f}_{\text{res}}, \boldsymbol{\tau}_{\text{res}}$.
When the compensation force is correctly used by the policy, the motion of the object matches the contact status between the hand and the object, and the residual force approaches zero.
At this point, the interaction motion should be physically reasonable.
Therefore, our physics reward $r^{\text{phys}}$ is computed as:
\begin{equation}
    r^{\text{phys}} = \text{exp}\left(-k^{\text{phys}}\left(\|\boldsymbol{f}_{\text{res}}\|+w^{\text{torque}}\|\boldsymbol{\tau}_{\text{res}}\|\right)\right).
\end{equation}
\section{Experiments}

In this section, we first introduce our implementation details and then compare our method with the previous state-of-the-art methods for HOI reconstruction. 
Next, we evaluate the key system design of our method. 
Finally, we discuss the limitations of our work. 
More video results can be found in our supplementary video.

\subsection{Implementation Details}
\begin{figure}[H]
    \centering
    \includegraphics[width=0.95\linewidth]{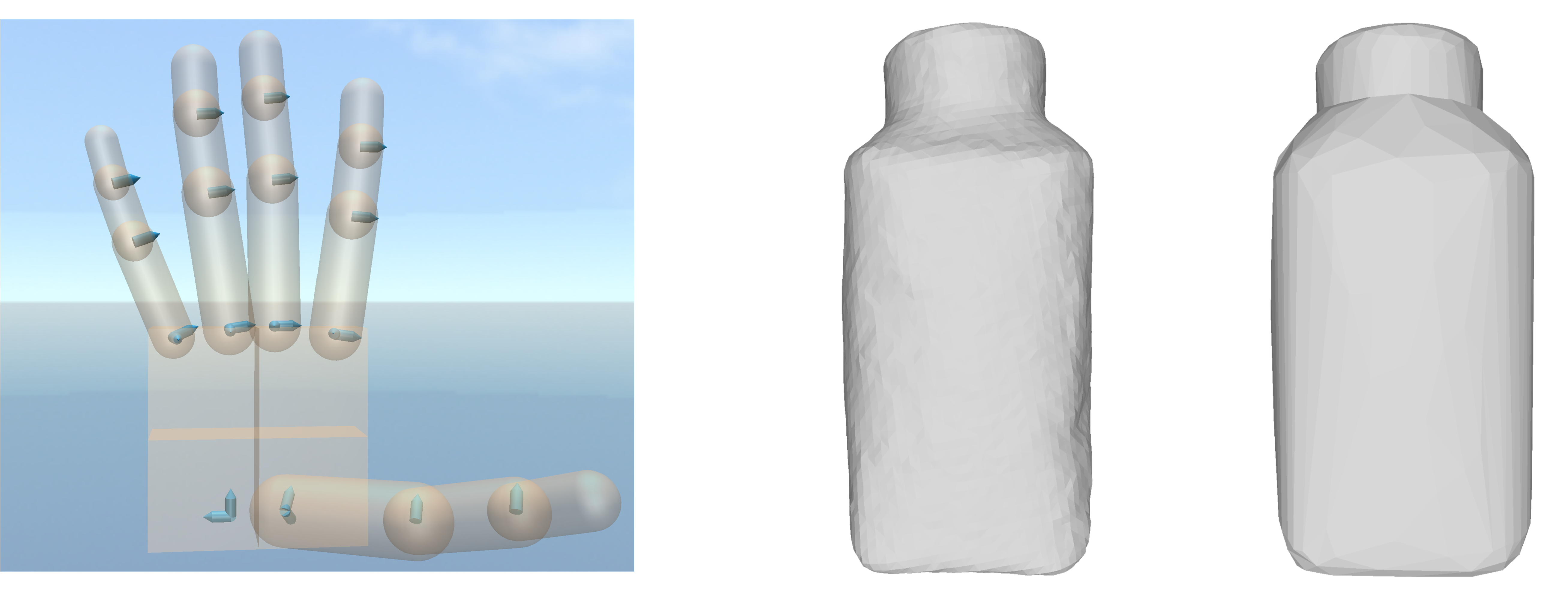}
    \caption{Our physical hand model (Left), object mesh reconstructed from kinematic tracking (Middle), and convex object collision mesh (Right).}
    \label{fig:phys_model}
\end{figure}

We utilize SingleDepth \cite{zhang2021single} as our kinematic estimator, which is the state-of-the-art method for real-time HOI reconstruction.
We use MuJoCo \cite{todorov2012mujoco} as our physics simulator.
Our physical hand model is shown in Fig.~\ref{fig:phys_model} (left), which has 26 DoFs and consists of capsules and boxes to represent the phalanges and the palm, respectively.
To obtain the object collision model (Fig.~\ref{fig:phys_model} (middle)), we first perform a convex decomposition algorithm \cite{wei2022coacd} on the object mesh (Fig.~\ref{fig:phys_model} (left)) reconstructed by our kinematic estimator.
Then we manually adjust the scale of the collision model to better approximate the reconstructed object mesh.
To achieve stable simulation, we utilize a frequency of 450Hz in the physics simulator, while the reference sequence operates at a frequency of 30Hz. 
This means that the simulator will step 15 times after the policy takes action once.
\par
To train the imitation policy, we collect a dataset using our kinematic estimator, which includes various hand-object interaction motions.
The dataset includes three objects: box, bottle, and banana, each with approximately 40 minutes (72K frames) of sequences.
For each object, we train a policy on the corresponding dataset.
During training, we follow the random start and early stop strategies proposed by \cite{peng2018deepmimic}.
In each episode, a motion clip is sampled from the dataset and the policy needs to imitate from a random start point till the sequence ends or a significant deviation occurs between the imitated results and the reference.
The training procedure takes about 2 days on a 16-core AMD EPYC 9654 CPU and an NVIDIA 4090 GPU, which consists of 200 million samples.
The inference speed of the policy is 8ms per frame, which enables us to combine it with our kinematic estimator without affecting the real-time performance.
Finally, our full tracking system can run at a speed of 25FPS with about 300ms delay.
The delay is majorly caused by using future kinematic results for the policy network $\pi$.
The number of future frames is controlled by $N$ in Eq.~\ref{eq:state}.
In practice, we set $N=5$, which means that the policy can take action with the knowledge of five future frames. 
Taking too many future frames does not significantly improve the policy (as shown in Tab.~\ref{tab:cmp-phys-plau}) and, at the same time, increases the system latency. 
Thus we set $N=5$ to make a trade-off.

\begin{table}
\caption{Average reward of the last 500 training epochs (the policy nearly converges) under different future frame number settings. Note that 1.0 is the maximum average reward that can be received by the policy.}
\label{tab:cmp-phys-plau}
\begin{tabular}{lccc}
\toprule
Number of future frames ($N$) & 1 & 5 & 10   \\
\midrule
Average Reward  &0.910 & \textbf{0.924} & 0.920          \\
\bottomrule
\end{tabular}
\end{table}

\begin{table}[]
\caption{Comparison results of tracking accuracy (APE, Object IoU) and physical plausibility (Phys., Pen., Hand Smo., Obj Smo.) with \cite{zhang2021single} and \cite{hu2022physical} on three interaction sequences. The unit of penetration is $mm$ while the unit of hand-object smoothness is $cm/s^2$. The non-zero penetration value of our method is due to the small difference between the collision object model and the reconstructed object model.
}
\label{tbl:cmp}
\resizebox{\linewidth}{!}{
\begin{tabular}{l|c|ccc}
\hline
                        & Method                 & {[}Zhang et al. 2021{]} & {[}Hu et al. 2022{]} & Ours  \\ \hline
\multirow{6}{*}{Box}    & APE $\downarrow$       & \textbf{12.75}          & 13.02                & 13.09          \\
                        & Obj IoU $\uparrow$     & 76.25                   & 74.93                & \textbf{78.14} \\
                        & Phys. $\uparrow$       & 59.56                   & 75.66                & \textbf{91.06} \\
                        & Pen. $\downarrow$      & 2.25                    & 2.33                 & \textbf{1.57}  \\
                        & Hand Smo. $\downarrow$ & 87.86                   & 97.99                & \textbf{70.28} \\
                        & Obj Smo. $\downarrow$  & 85.84                   & 88.03                & \textbf{70.72} \\ \hline
\multirow{6}{*}{Bottle} & APE $\downarrow$       & 12.18                  & 13.50                & \textbf{11.88} \\
                        & Obj IoU $\uparrow$     & 80.94                   & 80.19                & \textbf{81.12} \\
                        & Phys. $\uparrow$       & 73.12                   & 95.72                & \textbf{98.74} \\
                        & Pen. $\downarrow$      & 1.97                    & 1.83                 & \textbf{0.225} \\
                        & Hand Smo. $\downarrow$ & 131.3                   & 139.6                & \textbf{111.7} \\
                        & Obj Smo. $\downarrow$  & 129.6                   & 129.3                & \textbf{110.1} \\ \hline
\multirow{6}{*}{Banana} & APE $\downarrow$       & 10.53                   & 11.34                & \textbf{10.27} \\
                        & Obj IoU $\uparrow$     & 70.01                   & 73.26                & \textbf{73.53} \\
                        & Phys. $\uparrow$       & 85.86                   & 91.50                & \textbf{95.30} \\
                        & Pen. $\downarrow$      & 1.39                    & 1.63                 & \textbf{0.870} \\
                        & Hand Smo. $\downarrow$ & 68.28                   & 79.29                & \textbf{56.82} \\
                        & Obj Smo. $\downarrow$  & 69.67                   & 68.29                & \textbf{50.38} \\ \hline
\end{tabular}}
\end{table}

\subsection{Comparisons}
In the literature, the work of \cite{zhang2021single} and \cite{hu2022physical} are the state-of-the-art techniques of jointly tracking hand-object interactions in real-time with a single-view RGBD camera.
The former is used as our kinematic estimator while the latter, similar to us, aims to perform real-time physical optimization to better reconstruct the HOI motions.
To both evaluate the accuracy and the physical plausibility of the reconstructed interaction motions, we use the following metrics:
\begin{itemize}
\setlength{\leftmargin}{0pt}
    \item \textbf{Average Pixel Error (APE)} Following the previous work \cite{zhang2021single, hu2022physical}, we project the five fingertips of the tracked hand onto the input color image and evaluate the average pixel error to the ground truth.
    \item \textbf{Object IoU} To evaluate object tracking accuracy, we similarly project objects onto the image and measure the intersection over union (IoU) compared with the ground truth.
    \item \textbf{Physically Plausible Ratio (Phys.)} We evaluate the degree of consistency between the dynamic of the object and the hand-object contact status. 
    This is done with the same optimization approach as described in Sec. ~\ref{sec:surface}. When the sum of the residual force and torque after optimization is less than 0.01, we consider the frame to be physically plausible.
    \item \textbf{Penetration} To compute the penetration depth, we use collision models in the physics simulator to calculate the contact points between the hand and object for each frame, and then substitute these contact point positions into the SDF function of the reconstructed object mesh.
    \item \textbf{Hand and Object Smoothness (Hand Smo. \& Obj Smo.)} Similar to \cite{zhang2021manipnet}, we measure the smoothness of hand motion by the acceleration of the hand's joints and fingertips, and the smoothness of object motion by the acceleration of the object's center of mass.
\end{itemize}
\par
\begin{figure}[H]
    \centering
    \includegraphics[width=0.95\linewidth]{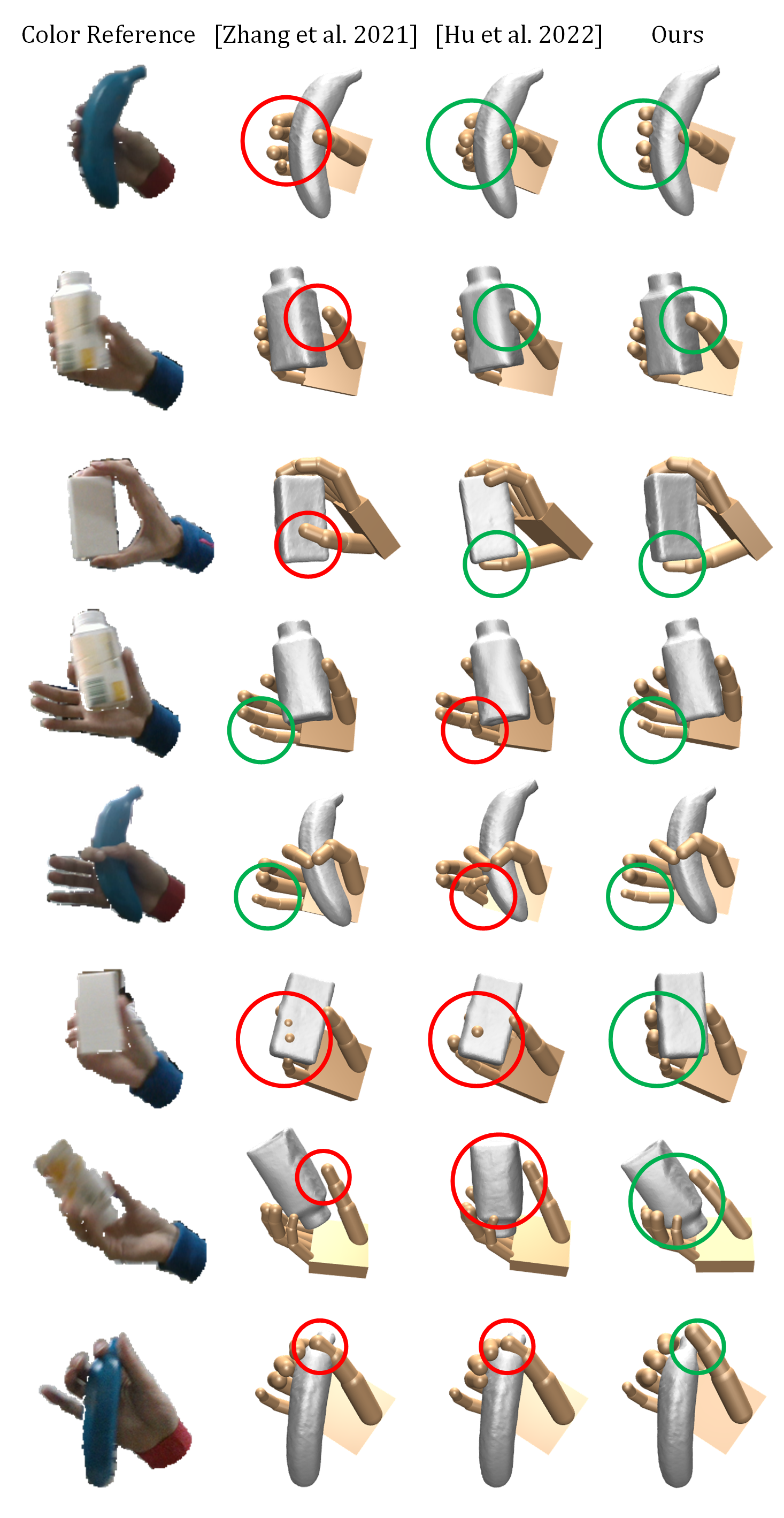}
    \caption{Qualitative comparison with \cite{zhang2021single} and \cite{hu2022physical}}
    \label{fig:quali}
\end{figure}

Our test set contains three labeled sequences with 5,417 frames in total, and each sequence corresponds to one object.
Tab.~\ref{tbl:cmp} shows the quantitative results of the three test sequences.
The results show that our method can achieve comparable tracking accuracy compared with our kinematic estimator \cite{zhang2021single} while significantly improving physical plausibility compared to both \cite{zhang2021single} and \cite{hu2022physical}.
Besides, the interaction motions reconstructed by our method are more smooth and with less penetration.
Fig.~\ref{fig:quali} illustrates some qualitative comparison results.
The first three rows in the figure show that our method, similar to \cite{hu2022physical}, can refine the missing contact points (R1, R2) and the wrong contact position caused by non-physical sliding (R3).
However, as \cite{hu2022physical} only models point contacts at the fingertips, it may erroneously optimize some contact points that do not exist (R4, R5), while our method avoids these artifacts by learning from motions with diverse contact status using a surface contact model.
Besides, our method exhibits higher robustness when facing rapid interactive movements (R6, R7). We believe this is due to the ability of physical controllers to perform smooth motions.
Finally, leveraging the non-penetration nature of physical simulation, our method effectively handles non-physical phenomena of hand-object penetration that occur in some kinematic tracking results (R8).
For more results, please refer to our video.

\subsection{Evaluation}
\begin{figure}[H]
    \centering
    \includegraphics[width=0.8\linewidth]{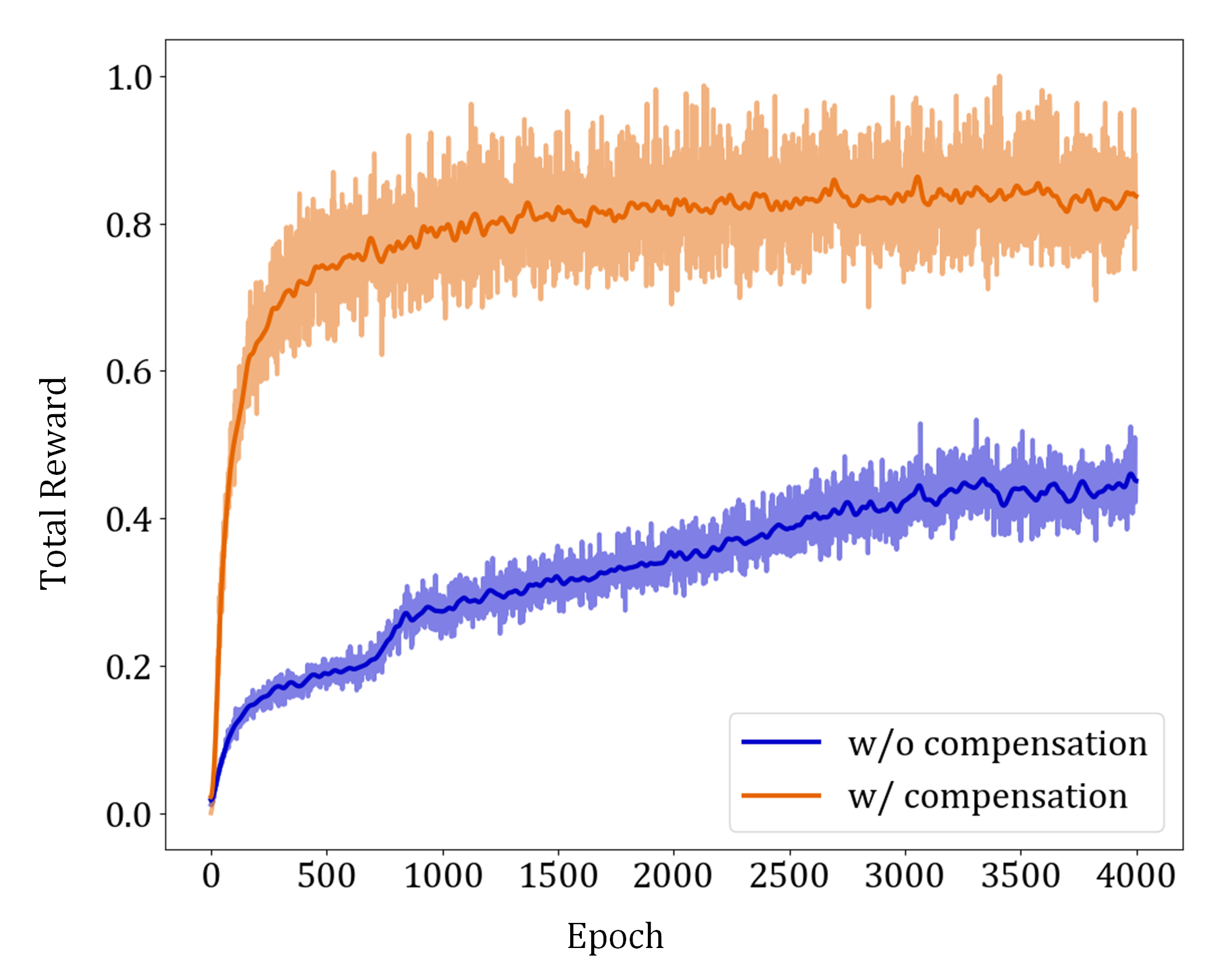}
    \caption{Training curve of our method with (orange) and without (blue) object compensation control. The Y-axis represents the total rewards obtained by the policy in all episodes within one epoch. The results are normalized by the maximum and minimum values of the total rewards across all epochs.}
    \label{fig:abl_compensation}
    \vspace{-1em}
\end{figure}

Firstly, we evaluate the effectiveness of the object compensation control.
Fig.~\ref{fig:abl_compensation} shows the training curve of our reinforcement learning framework w/ and w/o compensation control.
The result shows that the use of compensation control can significantly ease the learning process and improve the imitation quality.
We also demonstrate in the video that without the compensation control, the policy can only imitate some very simple interactive actions (such as static grasping), and cannot imitate more complex actions correctly.

\begin{figure}[H]
    \centering
    \includegraphics[width=0.95\linewidth]{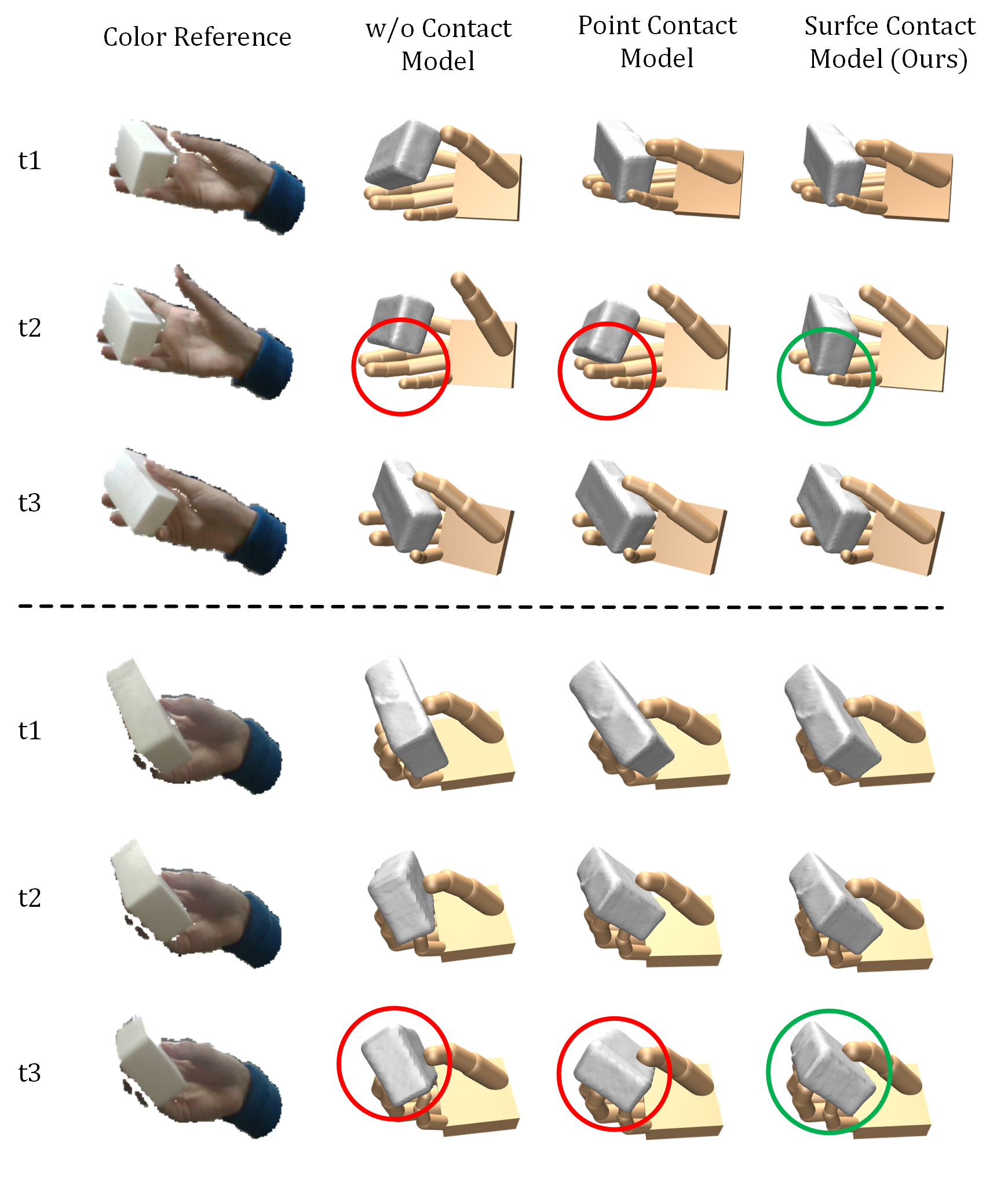}
    \caption{Qualitative evaluation of our surface contact model compared with two other designs: simply requiring minor compensation forces (w/o Contact Model), and a point contact model. 
    Our surface contact model offers more precise modeling of hand-object interaction, particularly in challenging scenarios involving rapidly changing contact status (Top). Additionally, it incorporates torsional and rolling friction torque, enabling more precise imitation of the rotational motion of objects (Bottom).}
    \label{fig:abl_scm}
\end{figure}

Then, we evaluate the efficacy of our surface contact model (SCM) by replacing it with two other designs: (1) completely remove the surface contact model and simply require minor compensation forces to form the physics reward $r_t^{\text{phys}}$ (w/o CM), (2) remove the extended contact points used to model surface contact and use a point contact model to compute physics reward (PCM). We compare the performance of these three approaches on the Box sequence. The quantitative results of reconstruction accuracy and physical correctness are shown in Tab.~\ref{tbl:abl_scm}, and the qualitative results are shown in Fig.~\ref{fig:abl_scm}.
The results indicate that when there is no contact model to explain the compensation force, the policy fails to learn to use the compensation force correctly, and thus more non-physical outcomes occur.
Although the point contact model provides a suitable approximation in most cases, it still falls short of accurately representing the actual contact mechanisms in some challenging interaction motions. In contrast, our surface contact model is closer to real hand-object interactions, therefore achieving better results.
For more results, please refer to our video.

\begin{table}[]
\caption{Quantitative comparison of our surface contact model (SCM) with point contact model (PCM) and no contact model to explain compensation force (w/o CM). The evaluation is performed on the Box sequence.}
\label{tbl:abl_scm}
\begin{tabular}{lccc}
\toprule
        & w/o CM & PCM & SCM (Ours) \\
\midrule
APE $\downarrow$     & 13.57             & 13.38               & \textbf{13.09}               \\
Obj IoU $\uparrow$ & 75.85             & 77.80               & \textbf{78.14}               \\
Phys. $\uparrow$   & 73.37             & 90.16               & \textbf{91.06}               \\
\bottomrule
\end{tabular}
\end{table}

\subsection{Limitations and Discussions}
The generalization capability of our system is not satisfying for the current technique as the system only works on each of the trained objects.
We believe this is due to the complexity of the HOI motion, which makes it very hard to learn the physics laws by the current deep reinforcement learning framework given the inaccurate physical simulator.
However, we still believe this is a step further in learning the physics of the complex HOI motion, and its benefits in reconstruction are well demonstrated.
From the numerical errors, the tracking accuracy of the hand and object judging from the observed view is not improved largely as the kinematic tracking method already achieves high performance on these metrics by directly optimizing them.
On the other hand, we see our method largely improves the physical plausibility while still maintaining comparable accuracy, which still indicates the advances of our technique.
Our technique is only evaluated with rigid objects with simple shapes.
it is still an open problem to handle more complex object shapes and motions.

\section{Conclusions}
We propose HOIC which learns the physics of hand-object interaction motions and improves the reconstruction of these motions in the case with limited recording (i.e. single-view RGBD recording). 
To the best of our knowledge, this is the first work that enables deep reinforcement learning to learn the physics of complex hand-object interactions.
The success of training the HOIC is attributed to a compensation force modeling method, which eases the learning process by directly manipulating the object, rather than using the hand joint torques to indirectly control the object. 
Furthermore, the non-real compensation force is also able to bridge the difference between the used point contact physical model and the real surface contact.
In this manner, the physical simulator does not need to model the soft hand tissues to generate surface contact, which is complex and time-consuming. 
Experiments demonstrated that our method successfully learns the complex physics of interaction motions and generates more physically plausible reconstruction results with limited observations.

\begin{acks}
This work was supported by the National Key R\&D Program of China (2023YFC3305600, 2018YFA0704000), the NSFC (No.62021002), and the Key Research and Development Project of Tibet Autonomous Region (XZ202101ZY0019G). This work was also supported by THUIBCS, Tsinghua University, and BLBCI, Beijing Municipal Education Commission.
\end{acks}

\bibliography{ref}

\clearpage

\appendix

\end{document}